\newcommand{\method}{{PubMed Reasoner}}
\documentclass[11pt]{article}

\usepackage[preprint]{acl}

\usepackage{times}
\usepackage{latexsym}

\usepackage[T1]{fontenc}
\usepackage[utf8]{inputenc}

\usepackage{microtype}

\usepackage{inconsolata}

\usepackage{graphicx}
\usepackage{amsmath}
\usepackage{algorithm}
\usepackage{algorithmic}
\usepackage{booktabs}
\usepackage{multirow}
\usepackage{listings}
\usepackage[listings, many]{tcolorbox}
\usepackage{amssymb}
\usepackage[utf8]{inputenc}
\usepackage[shortlabels]{enumitem}
\usepackage{mdframed,xcolor}
\newmdenv[linecolor=black,backgroundcolor=gray!6,skipabove=\baselineskip,skipbelow=\baselineskip]{vignette}
\usepackage{listings}
\lstdefinelanguage{none}{}

\title{\method{}: Dynamic Reasoning-based Retrieval for Evidence-Grounded Biomedical Question Answering}
\author{Yiqing Zhang\textsuperscript{1,2} \and Xiaozhong Liu\textsuperscript{2} \and
Fabricio Murai\textsuperscript{2}
\\
\\
 \textsuperscript{1}PayPal,
 \textsuperscript{2}Worcester Polytechnic Institute
\\
\\
 \texttt{ yiqizhang@paypal.com, \{xliu14,fmurai\}@wpi.edu}
  }

\begin{document}
\maketitle
\begin{abstract}
    Trustworthy biomedical question answering (QA) systems must not only provide accurate answers but also justify them with current, verifiable evidence. 
%Existing large language models (LLMs), however, rely on static parametric memory and can produce outdated or unsupported claims.
Retrieval-augmented approaches partially address this gap but lack mechanisms to iteratively refine poor queries, whereas self-reflection methods kick in only after full retrieval is completed. In this context, we introduce \method{}, a biomedical QA agent composed of three stages: \textbf{self-critic query refinement} evaluates MeSH terms for coverage, alignment, and redundancy to enhance PubMed queries based on partial (metadata) retrieval; \textbf{reflective retrieval} processes articles in batches until sufficient evidence is gathered; and \textbf{evidence-grounded response generation} produces answers with explicit citations. \method{} with a GPT-4o backbone achieves \textbf{78.32\%} accuracy on PubMedQA, slightly surpassing human experts, and showing consistent gains on MMLU Clinical Knowledge. Moreover, LLM-as-judge evaluations prefer our responses across: reasoning soundness, evidence grounding, clinical relevance, and trustworthiness. By orchestrating retrieval-first reasoning over authoritative sources, our approach provides practical assistance to clinicians and biomedical researchers while controlling compute and token costs.
\end{abstract}

\section{Introduction}

\begin{figure}[t]
\centering 
\includegraphics[width=\linewidth]{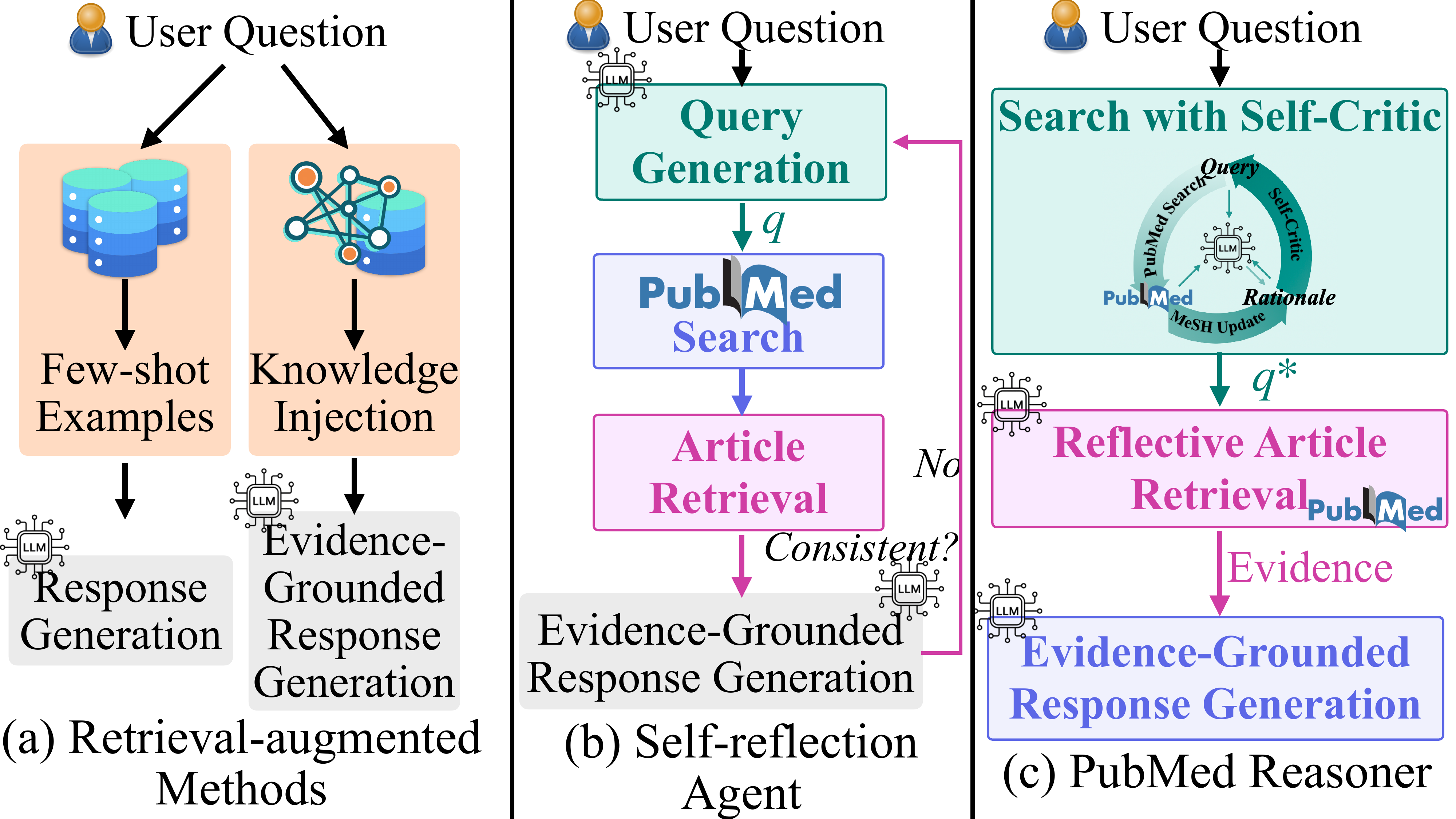}
\caption{
RAG, Self-reflection vs.\ \method{}. \textbf{(a) RAG baselines:} uses few-shot exemplars or custom databases but lack retrieval feedback. \textbf{(b) Self-reflection agents:} inspired our proposal; generates responses first and only reflects after completion.  \textbf{(c) \method{}:} a search-first approach that performs self-critic query refinement, reflective article retrieval in batches w/ early stopping once evidence is sufficient, and evidence-grounded response generation with explicit citations.
}
\label{fig:intro}
\end{figure}

Trustworthiness is essential in biomedical domains. Biomedical question answering (QA) systems must be factually grounded, current, and interpretable. Yet, large language models (LLMs) that rely primarily on parametric memory can hallucinate \cite{kalai2025language}, drift out of date, or omit key evidence \cite{guan2023language, xu2024hallucination}. While prior works have explored retrieval-augmented methods, to our knowledge, this is the first work to explicitly support citation-backed responses on biomedical QA datasets, ensuring that each response is transparently grounded in published evidence.

% \fm{In this para, I'm missing some works that are specifically trying to generate citation-supported answers with explanations for MedicalQA datasets. Are there any? If not, we have to mention this front and center.} This gap has driven much interest in retrieval-augmented generation (RAG). However, the two dominant RAG patterns face important limitations in biomedical settings:
Existing retrieval-augmented generation (RAG) approaches remain limited in biomedical settings (see Fig.~\ref{fig:intro}a):
\textbf{(i) Few-shot prompting} retrieves a few in-distribution exemplars (few-shot samples) and prompt the LLM to imitate them~\cite{nori2023can, nori2023capabilities}. While this can improve accuracy on similar cases, it does not yield structured explanations \cite{singhal2025toward}. The link between examples is built via clustering and similarity search, which requires sizable, well-labeled exemplar data. As a result, exemplar RAG is brittle and biased: it optimizes for local pattern matching rather than concept coverage or causal explanations. \textbf{(ii) Private knowledge databases} consist of bespoke stores such as entity graphs \cite{abu2024knowledge} or schema-aligned tables \cite{arslan2024survey} which can support step-wise explanations, but require strong priors and heavy maintenance, with limited reusability across domains. In biomedicine, where PubMed and MeSH (Medical Subject Headings) terms continue to grow, keeping such stores complete and current is prohibitively expensive.

A parallel line of work equips LLMs with web search capabilities (``deep research''), producing responses with citations. Yet, these systems typically lack the ability to constrain retrieval to authoritative biomedical sources such as PubMed, often yielding citations from less reliable or incomplete domains. 
Recent ``self-reflection'' agents (Figure~\ref{fig:intro}b) provide another direction, using consistency checks \cite{wang2022self} or reward-based signals \cite{leike2018scalable, shinn2023reflexion} to refine final answers. However, reflection occurs only after retrieval and response generation are complete, making the process computationally costly and unable to correct poor upstream retrieval or query formulation.

\noindent\textbf{Our approach.} 
We present \method{}, a multi-stage agent that mirrors the workflow of a biomedical researcher. 
Unlike prior methods that reflect only after producing responses, \method{} introduces feedback much earlier. During query planning, a self-critic evaluates candidate MeSH terms and their Boolean composition directly against live PubMed metadata, eliminating the need for static private databases. This structured feedback prevents low-quality queries from propagating downstream and balances recall with precision. The refined query is then issued to PubMed, where a reflective retriever processes articles in small batches and halts once sufficient evidence is gathered, controlling token usage. Finally, \method{} synthesizes an evidence-grounded response with explicit inline citations, ensuring transparency and interpretability.
%Instead of waiting until \fm{rephrase/clarify: response time}, \method{} uses a self-critic during query planning to evaluate candidate MeSH terms and the boolean expression that combines them directly against live PubMed metadata, without the need to build or access custom-made private databases. This early, structured feedback prevents low-quality queries from propagating downstream and optimizes recall without sacrificing precision. After issuing the refined PubMed query, a reflective retriever processes articles in small batches and early-stops once the evidence pool is sufficient, controlling token cost. Finally, \method{} composes an evidence-grounded response with explicit citations, ensuring traceability and interpretability.

Instead of treating the LLM as a one-shot generator, \method{} introduces a \textbf{dynamic reasoning-based paradigm}, orchestrating three interconnected stages over external evidence. 
%\fm{The phrasing is still strange in terms of framing components as contributions. I think the paradigm is an interesting contribution. Another can be the proposed system that showcases how to operationalize this paradigm in the biomedical QA context, then finally the evaluation of the system. We don't need to explain the components here.} 
This work makes the following contributions: 
 \setlist{nolistsep}
    \begin{itemize}[noitemsep]
\item We shift biomedical QA beyond one-shot generation by introducing an iterative workflow that plans, retrieves, and reasons over external evidence.
\item We integrate self-critic query refinement with batch-wise reflective retrieval over the PubMed database, enabling robust evidence grounding without maintaining private knowledge stores.  
\item We demonstrate consistent improvements in accuracy, explanation quality, and computational efficiency on PubMedQA and MMLU Clinical Knowledge, outperforming strong LLM, RAG and self-reflection baselines.

\end{itemize}

\section{Preliminaries}

\noindent\textbf{PubMed and MeSH Terms.} 
PubMed is the primary biomedical literature database, indexing over 35 million references. Each article is annotated with Medical Subject Headings (MeSH), a controlled vocabulary that organizes biomedical concepts into a hierarchical taxonomy. Queries often combine MeSH terms with Boolean operators (AND, OR, NOT), enabling structured retrieval.

\noindent\textbf{Biomedical Question Answering.}
The input is a natural language question $Q$ (e.g., “\emph{Do leukotrienes play a key role in asthma?}”), and the output is a response $R$ that is both factually accurate and explicitly supported by authoritative literature. 
%This task is challenging because biomedical concepts evolve rapidly, new studies are continuously added to PubMed, and retrieved evidence must be integrated into a clear, interpretable answer.

\noindent\textbf{Problem Setup.}
Given a user question $Q$, an optional task specification $T$  (e.g., ``\emph{answer yes/no with a justification}''), and optional context $C$, the objective is to retrieve a set of relevant biomedical articles $A$ and synthesize a final response $R$ that is evidence-grounded and interpretable. % Unlike prior RAG approaches that directly append retrieved documents to prompts, we structure the process as an iterative cycle of query refinement, evidence filtering, and response generation.

\noindent\textbf{Search Result Assessment Metrics.} \label{subsec: metrics}
To enable iterative query refinement, we adapt evaluation dimensions from search benchmarks \cite{gao2013bigdatabench, jiang2024mmsearch} to the biomedical QA setting, defining three structured feedback signals:
\begin{itemize}[leftmargin=8pt]
    \item \textbf{Coverage:} Does the MeSH term retrieve articles that represent the core biomedical concept?
    \item \textbf{Alignment:} Are the retrieved articles relevant to the original question?
    \item \textbf{Redundancy:} Does the MeSH term overlap with others, reducing retrieval efficiency?
\end{itemize}
These signals guide improvements to the evolving query, ensuring that downstream reasoning is grounded in a high-quality evidence pool.

\section{Proposed Method}
\begin{figure*}[t!] 
\centering 
\includegraphics[width=0.9\linewidth]{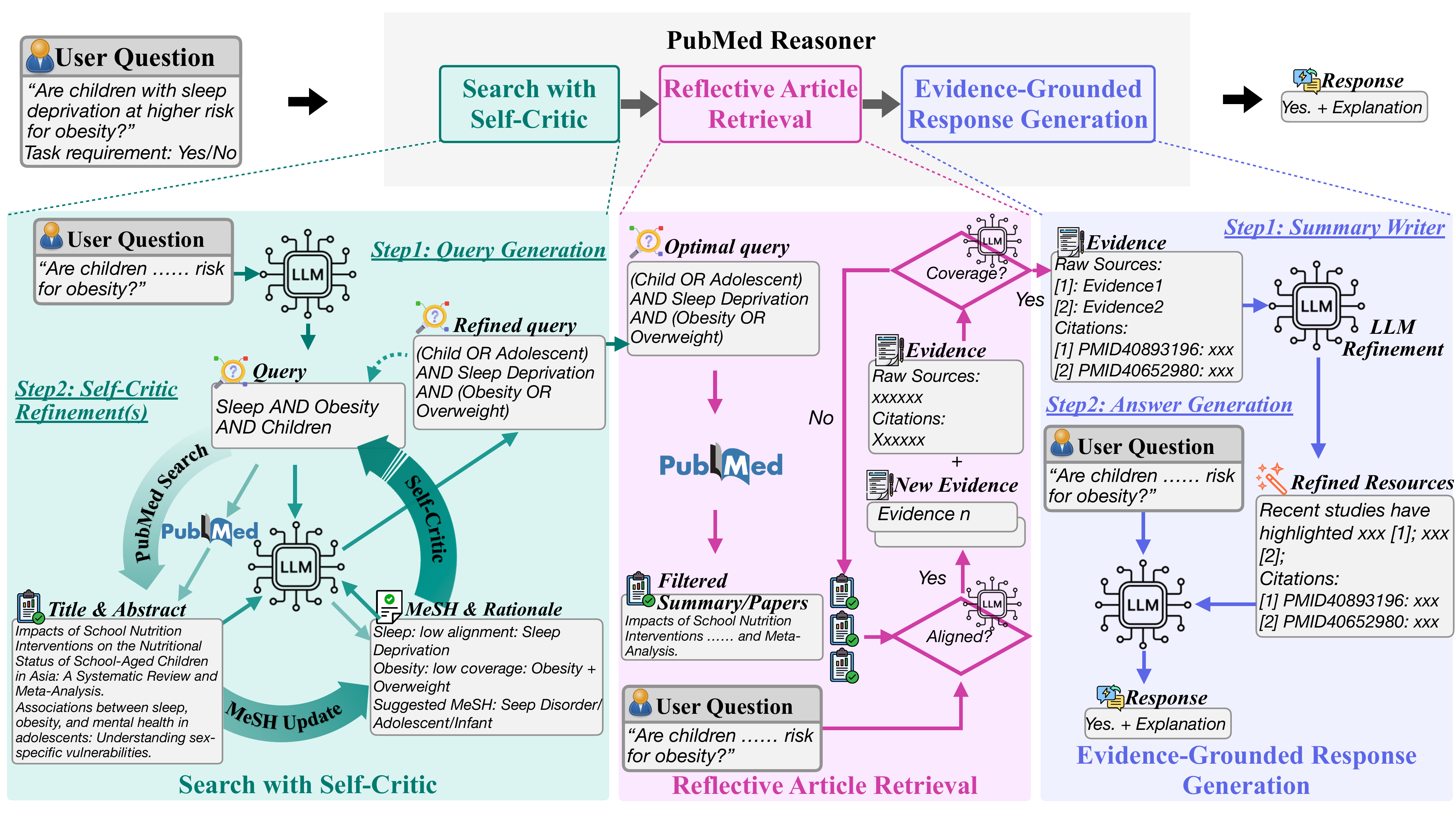}
\caption{
\method{} stages.
%has three interconnected stages that jointly optimize retrieval and reasoning. 
\textbf{(1) Search with Self-Critic Query Refinement.} From a user question, MeSH terms and a structured query are proposed. Self-critic evaluates each term for \emph{coverage}, \emph{alignment}, and \emph{redundancy}, iteratively refining the query. %toward an optimal query $q^{\ast}$. 
\textbf{(2) Reflective Article Retrieval with Early Stopping.} \method{} queries PubMed, filters results by title/abstract, and extracts supporting evidence in batches, checking whether accumulated evidence sufficiently covers the question; if so, retrieval is terminated early to save tokens and avoid unnecessary processing. 
\textbf{(3) Evidence-Grounded Response Generation.} Retained evidence is synthesized into final answer with explicit inline citations, ensuring factual grounding and traceability. %to original sources.
More case studies are provided in App.~\ref{app:case}.
}

% \vspace{-5mm}
\label{fig:model}
\end{figure*}

We introduce \method{}, a multi-stage agent framework inspired by the workflow of a biomedical researcher. Unlike direct LLM-based answering or one-shot retrieval-augmented generation, \method{} explicitly separates reasoning into three phases: 
\textbf{(1) Search with Self-Critic Query Refinement}, 
\textbf{(2) Reflective Article Retrieval with Early Stopping}, and 
\textbf{(3) Evidence-Grounded Response Generation}. 
This design ensures that the system does not rely solely on parametric memory, but grounds its explanation in authoritative biomedical literature.

\subsection{Search with Self-Critic Query Refinement}
\textbf{Query Generation.}
The first stage constructs a structured query that serves as the initial search plan for PubMed retrieval. Given a user question \( Q \) grounded in biomedical knowledge, together with optional contextual information \( C \), a large language model (LLM) is employed to generate a broad set of candidate Medical Subject Headings (MeSH) terms. Each candidate term is accompanied by a brief rationale explaining its relevance to the query. Formally, this process is defined as:
\begin{equation}
\mathcal{M}_0 \leftarrow \mathrm{LLM}_{\mathrm{mesh\_gen}}(Q, C),
\end{equation}
where \( \mathcal{M}_0 = \{ (\mathrm{MeSH}_i, r_i) \}_{i=1}^{N} \), \( \mathrm{MeSH}_i \) denotes a candidate MeSH term, and \( r_i \) provides the corresponding semantic justification for its inclusion.

From the initial candidate pool $\mathcal{M}_0$, the LLM selects a subset of MeSH terms based on multiple criteria, including the model’s confidence for each term, the plausibility of its accompanying rationale, and the degree of semantic alignment between the term and the input question:
\begin{equation}
\mathcal{M}_0' = \mathrm{LLM}_{\mathrm{mesh\_select}}(\mathcal{M}_0, Q, C),
\end{equation}
where $\mathcal{M}_0' \subseteq \mathcal{M}_0$ denotes the set of candidate terms with their rationales. 
The selected subset $\mathcal{M}_0'$ is then combined using Boolean operators to form the initial structured query:
\begin{equation}
q_0 \leftarrow \mathrm{LLM}_{\mathrm{query\_gen}}(\mathcal{M}_0', Q, C).
\label{equ:query_gen}
\end{equation}
Core concepts are typically linked with AND to enforce precision, while broader synonyms or alternatives are connected with OR to maximize recall. Temporal filters (e.g., publication date restrictions) are then applied to constrain retrieval during evaluation. This formulation mirrors the workflow of human researchers, who begin with broad but structured queries and iteratively refine them based on preliminary search results.

\noindent\textbf{Iterative Self-Critic Query Refinement.} Once the initial query is generated, the self-critic mechanism guides iterative refinement. At each iteration $t$, \method{} submits the current query $q_{t-1}$ to the PubMed search engine, which returns a ranked list of retrieved records (e.g., title, abstract, and PubMed ID):
\begin{equation}
\mathcal{A}_t \leftarrow \mathrm{PubMedSearch}(q_{t-1}).
\label{equ:search}
\end{equation}

From the ranked results $\mathcal{A}_t$, we extract metadata fields that serve as inputs to the self-critic. We denote these self-critic signals as:
\begin{equation}
\mathcal{S}_t = \big\{ (\text{title}_i,\, \text{abstract}_i) \big\}_{i=1}^{|\mathcal{A}_t|}.
\label{equ:metadata}
\end{equation}

Rather than analyzing full texts which would incur substantial computational and token costs, the self-critic operates solely on \(\mathcal{S}_t\). This design enables efficient evaluation of candidate MeSH terms while preserving sufficient semantic information for relevance assessment. Given the current candidate MeSH pool \(\mathcal{M}_{t-1}'\), each term is then evaluated along three structured dimensions (Sec.~\ref{subsec: metrics}), which collectively provide actionable feedback for iterative query refinement: 
\underline{Coverage} measures whether the concept represented by a candidate $\text{MeSH}_i$ term appears in the self-critic signals $\mathcal{S}_t$; 
\underline{Alignment} evaluates whether the articles associated with each  candidate $\text{MeSH}_i$ term are pertinent to the user question; 
\underline{Redundancy} identifies whether a candidate term overlaps with, or is superfluous given other terms in the current set, as well as the logical composition (AND/OR) implied by the query intent.

The evaluation of \(\mathcal{M}_{t-1}'\) along these dimensions is done through the following operators:
\begin{align}
\mathrm{Cvg}_t & \leftarrow \mathrm{LLM}_{\mathrm{coverage}}\big(\mathcal{M}_{t-1}', \mathcal{S}_t \big),
\label{equ:coverage} \\
\mathrm{Align}_t & \leftarrow
\mathrm{LLM}_{\mathrm{alignment}}\big(\mathcal{M}_{t-1}', \mathcal{S}_t, Q\big),
\label{equ:alignment} \\
    \mathrm{Redun}_t & \leftarrow \mathrm{LLM}_{\mathrm{redundancy}}\big(\mathcal{M}_{t-1}', \mathcal{S}_t, q_{t-1} \big).
\label{equ:redundancy}
\end{align}
Each of $\mathrm{Cvg}_t$, $\mathrm{Align}_t$, and $\mathrm{Redun}_t$ is a list of pairs $\{(y_{t,i}, \ r_{t,i})\}^{|\mathcal{M}_{t-1}'|}_{i=1}$, where $y_{t,i} \in \{\text{Yes}, \text{No}\}$ is a binary outcome indicating the result of the corresponding operator for candidate term $\text{MeSH}_i$, and $r_{t,i}$ provides a textual rationale that informs subsequent refinement. In particular, for $\mathrm{Redun}_t$, $r_{t,i}$ both explains the redundancy verdict and includes the recommended boolean linkage under the previous search query $q_{t-1}$.

Given the previous \text{MeSH} terms, current self-critic signals $\mathcal{S}_t$, and per–term feedback, we refine the candidate MeSH set:
\begin{equation}
\begin{split}
    \mathcal{M}_t\ \leftarrow\ 
    \mathrm{LLM} &_{\text{update}} \!\Big(
    \ \mathcal{M}_{t-1}', \  Q,\ C,\ \\ \mathcal{S}_t, \ 
    &{\text{Cvg}}_t,\ {\text{Align}}_t,\ {\text{Redun}}_t 
\Big),
\end{split}
\label{equ:update}
\end{equation}
where $\mathcal{M}_t$ is the revised set (with rationales). The update favors terms that increase coverage without harming alignment, and prunes (or merges) terms flagged as redundant.

From the refined \text{MeSH} terms $\mathcal{M}_t$, \method{} drafts a new query, then enforces syntactic validity with a rule-based normalizer:
\begin{align}
\tilde{q}_t\ &\leftarrow\
\mathrm{LLM}_{\mathrm{refine}}\big(\mathcal{M}_t,\ \mathcal{S}_t,\ \mathcal{H},\ Q,\ C\big),
\label{equ:refine-draft}
\\
q_t\ &\leftarrow\
\mathrm{QueryNormalize}\big(\tilde{q}_t\big),
\label{equ:refine-validate}
\end{align}
where $\mathcal{H}$ stores the query history $\{q_0 ,..., q_{t-1}\}$. The refinement aims to balance recall and precision using the proxies above: encourage additions that improve coverage (recall) if alignment remains high, and remove or demote terms that are misaligned or redundant (precision). The normalizer applies rule-based checks for operator placement, parenthesization, and field qualifiers, repairing common errors to guarantee a valid PubMed query. The self-critic loop then repeats with $q_t$ until a stopping rule is met (e.g., no gain in coverage/alignment or a budget limit). The final optimized query $q^\ast$ is returned for downstream retrieval. 
A detailed algorithm is provided in Algo.~\ref{algo:self-critic}, and the prompts for each stage are included in App.~\ref{app:prompts}.

\subsection{Reflective Article Retrieval with\ \ \ \ \ \ \ \ \ Early Stopping}
\method{} operates on the final retrieved records
$\mathcal{A}^\star=\{a_1,\dots,a_M\}$ returned by the search step (Eq.~\ref{equ:search}) together with their corresponding metadata $\mathcal{S}^*$(Eq.~\ref{equ:metadata}).
Since PubMed search prioritizes records by relevance, highly pertinent evidence is expected to appear early in the ranked list. Accordingly,
\method{} enforces an early-stopping rule: once sufficient supporting evidence has been accumulated, retrieval halts to avoid further token consumption.

\noindent\textbf{Coarse Filtering.}
Each retrieved record is first screened using metadata $\mathcal{S}^*$ to assess coarse relevance to the query. Only plausibly relevant records are retained for subsequent processing:
\begin{equation}
\big\{\big(v_i, r^{\text{filter}}_i\big)\big\}_{i=1}^{|\mathcal{A}^\star|} \leftarrow 
\mathrm{LLM}_{\mathrm{filter}}\big(
\mathcal{S}^\star, \ Q
\big),
\label{equ:filter}
\end{equation}
where $v_i\!\in\!\{\text{Yes},\text{No}\}$ indicates keep/drop and $r^{\text{filter}}_i$ provides a short rationale. Since $\mathcal{A}^\star$ is ranked by PubMed, the retained set 
$\mathcal{A}^{+}=\{\,a_i\in\mathcal{A}^\star:\ v_i=\text{Yes}\,\}$ 
preserves the original order and enables early prioritization of high-quality evidence.
To control retrieval cost, coarse filtering is applied only to the top $M_{\max}$ ranked articles, the maximum budget of articles allowed for downstream processing.
Additional details regarding $M_{\max}$ and token budget $T$ are provided in App.~\ref{appd:settings}.

\noindent\textbf{Reflective Evidence Extraction.}
For each $a_i\in\mathcal{A}^{+}$, we extract candidate evidence and evaluate \textbf{alignment} (i.e., does the candidate evidence directly address $Q$?):
\begin{equation}
\big\{(E_i,\ \mathrm{align}_i,\ r^{\mathrm{align}}_i)\big\}_{i=1}^{|\mathcal{A}^{+}|}
= \mathrm{LLM}_{\mathrm{extract}}\!\big(\mathcal{A}^{+}, Q\big),
\label{equ:extract}
\end{equation}
where $E_i$ is the extracted passage, $\mathrm{align}_i\!\in\!\{\text{Yes},\text{No}\}$ indicates whether the passage directly addresses $Q$, and $r^{\mathrm{align}}_i$ provides a rationale. 
The evolving evidence pool is then
\begin{equation}
\mathcal{E} \;=\; \{\,E_i : \mathrm{align}_i=\text{Yes}\,\}.
\label{equ:pool}
\end{equation}

% and admit an evidence only if and only if both signals are positive:
% \begin{equation}
% \mathcal{E}
% \;=\;
% \big\{\,E_i\ :\
% \mathrm{cvg}^{\mathrm{a}}_i=\text{Yes}\ \wedge\
% \mathrm{align}^{\mathrm{a}}_i=\text{Yes}
% \big\}.
% \label{equ:findings}
% \end{equation}
%
% Here $\mathcal{E}$ is the \fm{evolving???} pool of extracted, question-relevant evidence (chunks) with corresponding citations, aggregated across admitted articles.

\noindent\textbf{Batching and Early Stopping.} \label{sec:early-stop}
To reduce token costs, we process the ranked results in batches. We partition $\mathcal{A}^{+}$ into batches
$\{\mathcal{B}_1,\ldots,\mathcal{B}_K\}$ of size $m$ (e.g., $m{=}5$). After processing batch $b$, we update the evidence pool:
$
\mathcal{E}^{(b)}=\mathcal{E}^{(b-1)}\cup\{E_i: a_i\in\mathcal{B}_b, \ \mathrm{align}_i=\text{Yes}\},
\quad \mathcal{E}^{(0)}=\varnothing.
$
A reflective module then judges whether the current evidence $\mathcal{E}^{(b)}$ provides sufficient \textbf{coverage} of the query (i.e., whether the major aspects of $Q$ are addressed). 
\begin{equation}
    \texttt{is\_sufficient} \leftarrow \mathrm{LLM_{CoverageCheck}}(\mathcal{E}^{(b)}, Q)
\label{equ:coveragecheck}
\end{equation}
Retrieval stops early when $\texttt{is\_sufficient} = \text{Yes}$, the marginal utility of additional articles becomes negligible, or the token budget is reached. A detailed algorithm is provided in Algo.~\ref{algo:reflective}, and the prompts for each stage are included in App.~\ref{app:prompts}.

\subsection{Evidence-Grounded Response Generation}

Given the final curated evidence pool $\mathcal{E}$, \method{} composes a final response by integrating the most relevant findings into a coherent, citation-supported explanation.

\noindent\textbf{Summary–of–Evidence (SoE).}
To convert the vetted evidence $\mathcal{E}$ into a compact and citable representation, \method{} groups passages by article and distills key observations that directly address question $Q$. This produces a structured summary
$\mathrm{SoE} \leftarrow \mathrm{LLM}_{\mathrm{summary}}(\mathcal{E}, Q)$,
ensuring that every factual claim remains linked to its supporting source. Each retained citation is preserved explicitly, promoting transparency and reproducibility.

\noindent\textbf{Response Generation.}
Finally, the system generates the user-facing answer by conditioning on the question $Q$, task requirements $T$, optional context $C$, and the SoE. Formally, the final response is produced as
$R \leftarrow \mathrm{LLM}_{\mathrm{response}}(Q, T, C, \mathrm{SoE})$,
ensuring that the explanation is grounded, verifiable, and aligned with the task specification.

This staged design turns the LLM into a reasoning orchestrator: every statement in
the final answer is linked to specific citations, improving interpretability and clinical trustworthiness. Implementation details are provided in App.~\ref{appd:settings}.

\section{Experiments}

We evaluate \method{} on two biomedical QA datasets: PubMedQA~\cite{jin2019pubmedqa} and MMLU Clinical Knowledge (MMLU-CK)~\cite{wang2024mmlu}. For each dataset, we report both prediction accuracy and explanation quality, and conduct ablation studies to quantify the contribution of each component of \method{} to overall performance. PubMedQA additionally provides supporting context for each question; in our setting, we restrict retrieval to articles published within the official dataset’s specified year range to ensure consistency.

\subsection{Evaluation Metrics} \label{eval}
We assess model performance along several dimensions: prediction accuracy, explanation quality, cost analysis and evidence sufficiency depth.

\begin{itemize}[leftmargin=8pt,noitemsep]
    \item \textbf{Accuracy:} Proportion of questions for which the model produces the correct answer label.
 \item \textbf{Explanation Quality:} 
Using an LLM-as-a-judge (see prompt in App.~\ref{app:llm-judge}), we perform pairwise comparisons and assign five-point Likert scores on four axes: reasoning soundness, evidence grounding, clinical relevance and trustworthiness.
\item \textbf{Cost Analysis:}
To quantify the computational efficiency of each method, we measure four complementary cost indicators: (1) input token usage, (2) output token usage, (3) number of LLM API calls, and (4) number of PubMed search calls issued during retrieval. 
\item \textbf{Query Term Quality:}
To evaluate the effectiveness of search query formulation, we measure precision and recall of the MeSH terms generated after self-critic refinement, both crucial for downstream retrieval and reasoning performance.

\item \textbf{Evidence Sufficiency Depth (ESD):}
We additionally track how many articles are processed before early stopping is triggered. This metric reflects retrieval efficiency. Detailed results are provided in App.~\ref{app:esd}.
\end{itemize}
Detailed definitions and justification for each metric are provided in App.~\ref{sec:evalmetrics}.

\noindent

\subsection{Baselines}
We compare \method{} with the following representative baselines:
\begin{itemize}[leftmargin=8pt,noitemsep]
\item\textbf{LLM Baseline:} A strong LLM that answers questions without explicit retrieval or search planning.
\item\textbf{Human Performance:} Reported accuracy on PubMedQA, which serves as an approximate upper bound for model performance.
\item\textbf{RAG Method:} LLM generates a single search query, retrieves relevant articles from PubMed search engine, and incorporates them into response generation.

\item\textbf{Self-reflection Agent:} A reasoning-based baseline that combines retrieval with self-reflection to gradually refine the final answer. 
\end{itemize}
Detailed configuration of each baseline can be found in App.~\ref{appd:baseline}.

\subsection{PubMedQA}

\begin{table}[t]
\small
\centering
\caption{
Accuracy on PubMedQA and MMLU-CK test sets for \method{} variants, strong LLMs, RAG method, self-reflection agents, and human performance. 
Best result per column is \textbf{bold-faced}; second best is \underline{underlined}. Additional results for \method{} with Qwen variants are provided in App.~\ref{app:qwen-results}.
}
\label{tab:main_results}
\setlength\tabcolsep{3pt}
\begin{tabular}{llcc}
\toprule
\multicolumn{2}{c}{Method} & PubMedQA & MMLU-CK \\
\midrule
\multirow{3}{*}{\it \shortstack[l]{Without\\retrieval}} 
    & Gemini-2.5 Pro & 75.64\% & 60.52\% \\
    & GPT4o {\footnotesize (leaderboard)} & 75.20\% & 60.64\% \\
    & Human performance & 78.00\% & -- \\
\midrule
\multirow{2}{*}{\it \shortstack[l]{RAG\\method}} 
    & with Gemini & 72.30\% & 58.94\% \\
    & with GPT & 73.28\% & 58.26\% \\
\midrule
\multirow{2}{*}{\it \shortstack[l]{Self-\\reflection}} 
    & with Gemini & 77.08\% & 60.67\% \\
    & with GPT & 77.12\% & 60.79\% \\
\midrule
\multirow{2}{*}{\it \shortstack[l]{PubMed\\Reasoner}} 
    & with Gemini & \underline{77.26\%} & \underline{61.36\%} \\
    & with GPT & \textbf{78.32\%} & \textbf{63.21\%} \\
\bottomrule
\end{tabular}
\vspace{-4pt}
\end{table}

\begin{table*}[t!]
\caption{
Explanation quality evaluation on PubMedQA (left) and MMLU-CK (right) test sets: Gemini w/o retrieval vs.\ \method{}. % (\textit{Ours}).
Win/tie/loss rates from pairwise comparisons judged by GPT-4; average Likert scores (1–5).
% We compare Gemini without retrieval and PubMed Reasoner (\textit{Ours}).
}
\centering
\small
\setlength{\tabcolsep}{2pt}
\begin{tabular}{l ccc cc ccc cc}
\toprule
\multirow{3}{*}{Metric} & \multicolumn{5}{c}{PubMedQA} & \multicolumn{5}{c}{MMLU-CK} \\
\cmidrule(lr){2-6} \cmidrule(lr){7-11}
& \multicolumn{3}{c}{Loss / Tie / Win (\%)} & \multicolumn{2}{c}{Avg. Likert (1–5)} & \multicolumn{3}{c}{Loss / Tie / Win (\%)} & \multicolumn{2}{c}{Avg. Likert (1–5)} \\
\cmidrule(lr){2-4} \cmidrule(lr){5-6} \cmidrule(lr){7-9} \cmidrule(lr){10-11}
& Gemini & Tie & Ours & Gemini & Ours & Gemini & Tie & Ours & Gemini & Ours \\
\midrule
Reasoning Soundness & 39.7 & 5.7 & \textbf{54.6} & 3.416 & \textbf{3.584} & 44.0 & 10.8 & \textbf{45.2} & 3.307 & \textbf{3.699} \\
Evidence Grounding  & 40.8 & 3.8 & \textbf{55.4} & 3.421 & \textbf{3.601} & 25.2 & 11.7 & \textbf{64.1} & 3.209 & \textbf{3.595} \\
Clinical Relevance  & 39.2 & 7.4 & \textbf{53.4} & 3.438 & \textbf{3.584} & 34.4 & 20.0 & \textbf{45.6} & 3.525 & \textbf{3.732} \\
Trustworthiness     & 38.7 & 5.4 & \textbf{55.9} & 3.424 & \textbf{3.587} & 35.3 & 15.9 & \textbf{48.8} & 3.386 & \textbf{3.712} \\
\bottomrule
\end{tabular}
\label{tab:combined_eval}
\end{table*}

\noindent\textbf{Accuracy.}
Table~\ref{tab:main_results} shows that \method{} achieves near-human or superior accuracy on PubMedQA. In particular, \method{} (GPT) attains 78.32\% accuracy, slightly exceeding the reported human expert performance. Compared to other baselines, \method{} consistently outperforms both direct LLM inference and RAG-based methods, improving upon the GPT baseline by 3.12\% and yielding clear gains over RAG. Notably, standard RAG underperforms direct LLM inference, likely due to noisy retrieval caused by unrefined, one-shot query generation that introduces weakly aligned or irrelevant evidence into the model context. Relative to the self-reflection agent, \method{} achieves modest but consistent accuracy improvements.
Importantly, these gains are obtained with substantially lower computational cost (shown in Table~\ref{tab:cost}). 
%As shown in Table~\ref{tab:cost}, \method{} reduces input token usage, the number of LLM calls, and PubMed search calls by 29--55\% compared to the self-reflection baseline. 
Together, these results demonstrate that \method{} offers a more favorable accuracy--efficiency trade-off than existing baselines, delivering practical accuracy improvements while significantly reducing computational overhead. Qualitative error analysis and representative failure cases are provided in App.~\ref{app:error-analysis}.

\noindent\textbf{Reasoning Quality.} On PubMedQA, \method{} with the Gemini backbone consistently outperforms Gemini-2.5 Pro across all four LLM-judge dimensions, as reported in Table~\ref{tab:combined_eval}. The pairwise win rate rises by 14.9\% in Reasoning Soundness, by 14.6\% in Evidence Grounding, by 14.2\% in Clinical Relevance, and by 17.2\% in Trustworthiness, while tie rates remain low. The average Likert score also increases respectively by 0.168, 0.180, 0.146, and 0.163, yielding explanations that are more coherent, better grounded in evidence, more clinically focused, and more trustworthy.

\begin{table*}[t!]
\setlength{\tabcolsep}{4pt}
\caption{Cost comparison on PubMedQA and MMLU-CK using a shared GPT-4 backbone for \method{} and baselines. 
For PubMedQA, input and output token counts are reported as ratios relative to the direct LLM baseline. 
For each metric, the lower value between the self-reflection agent and \method{} is \textbf{bolded}.}
\label{tab:cost}
\small
\centering
\begin{tabular}{lrrrrrrrrr}
\toprule
 & \multicolumn{4}{c}{PubMedQA} && \multicolumn{4}{c}{MMLU-CK}\\ \cmidrule(lr){2-5}\cmidrule(lr){7-10}
 & \textbf{Input} & \textbf{Output} & \textbf{LLM} & \textbf{Search}  && \textbf{Input} & \textbf{Output} & \textbf{LLM} & \textbf{Search}\\
\textbf{Method} & \textbf{Tokens} & \textbf{Tokens} & \textbf{Calls} & \textbf{Calls} && \textbf{Tokens} & \textbf{Tokens} & \textbf{Calls} & \textbf{Calls}\\
\midrule
LLM     & 542.70      & 144.90      & 1     & 0  && 90.24     & 98.64      & 1     & 0  \\
RAG method     & $\times$1.64      & $\times$2.40     & 2     & 1  && $\times$4.00     & $\times$3.06      & 2     & 1  \\
Self-reflection    & $\times$225.27 & $\times$13.60 & 13.08 & 3.52 && $\times$1914.53 & \textbf{$\times$24.35} & 14.61 & 4.77\\
\method{} & \textbf{$\times$97.25}  & \textbf{$\times$12.67} & \textbf{7.61}  & \textbf{2.49} && \textbf{$\times$1098.26}  & $\times$24.86 & \textbf{10.72}  & \textbf{3.59}\\
\bottomrule
\end{tabular}
\end{table*}

\noindent \textbf{Cost Analysis.}
As shown in Table~\ref{tab:cost}, \method{} achieves substantial reductions in computational overhead compared to the self-reflection baseline. Across key cost metrics, \method{} reduces input token usage by 55.34\%, lowers the number of LLM API calls by 41.82\%, and decreases PubMed search calls by 41.36\%. These efficiency gains stem from structured query planning and early stopping, which avoid unnecessary retrieval and redundant reasoning steps. Although the overall cost of \method{} remains higher than that of the one-shot LLM baseline, it produces citation-grounded and verifiable responses, offering substantially stronger reliability than direct generation. Compared to the RAG baseline, which often yields shallow or weakly aligned search results, \method{} performs more targeted retrieval and accumulates higher-quality evidence.

\begin{table}[t]
\caption{Search quality on PubMedQA. Precision/recall computed by comparing set of predicted MeSH terms from the final query against gold MeSH annotations.}
\small
\centering
\setlength{\tabcolsep}{8pt}
\begin{tabular}{lcc}
\toprule
Method & Precision & Recall \\
\midrule
Gemini-2.5 Pro      & 0.9422 & 0.6848 \\
~+Self-reflection          & 0.9745 & 0.7900 \\
~+\method{}                  & \textbf{0.9826} & \textbf{0.8025} \\
\midrule
GPT-4o         & \textbf{0.9874} & 0.7052 \\
~+Self-reflection       & 0.9562 & 0.7928 \\
~+\method{}                     & 0.9868 & \textbf{0.8532} \\
\bottomrule
\end{tabular}
\label{tab:pubmedqa_search}
\end{table}

\noindent\textbf{Query Term Quality.} As shown in Table~\ref{tab:pubmedqa_search}, \method{} maintains high precision while substantially improving recall due to self-critic refinement during query planning. With a GPT backbone, recall improves by 20.99\% over the GPT baseline and by 7.6\% over the self-reflection agent. With a Gemini backbone, recall improves by 17.18\% over the baseline and also surpasses the self-reflection agent. These results demonstrate that self-critic refinement broadens conceptual coverage without sacrificing alignment, producing higher-quality search queries.

\subsection{MMLU Clinical Knowledge}

\noindent\textbf{Accuracy.} On MMLU-CK dataset (Table~\ref{tab:main_results}), self-reflection offers only a marginal lift over the raw LLM baseline. In contrast, \method{} delivers consistent gains over the respective backbones: 1.11\% over Gemini-2.5 Pro and 2.69\% over GPT-4o. This underscores that our self-critic improves accuracy beyond what self-reflection can achieve. Moreover, as shown in Table~\ref{tab:cost}, these accuracy gains are achieved with substantially lower computational overhead.%, reinforcing the efficiency of our approach.

\noindent\textbf{Explanation Quality.} On MMLU-CK, \method{} consistently surpasses Gemini-2.5 Pro across all LLM-judge dimensions (Table~\ref{tab:combined_eval}). Pairwise win rates rise by 1.2\% in reasoning soundness, 38.9\% in evidence grounding, 11.2\% in clinical relevance, and 13.5\% in trustworthiness, with low tie rates. Average Likert scores also increase by 6–12\% across dimensions, indicating clearer logic, stronger evidence support, sharper clinical focus, and greater trustworthiness. In sum, \method{} produces better clinical explanations. 

\noindent \textbf{Cost Analysis.} As shown in Table~\ref{tab:cost}, the MMLU-CK results exhibit a pattern similar to PubMedQA: \method{} consistently incurs substantially lower computational cost than the self-reflection baseline. 
In particular, \method{} requires fewer input tokens, fewer LLM API calls, and fewer PubMed search calls, reflecting a more efficient retrieval and reasoning workflow on this broader clinical knowledge benchmark as well.

\noindent \textbf{Query Term Quality.} MMLU does not provide MeSH term annotations, hence cannot be assessed.

\subsection{Ablation Study}
\noindent\textbf{Ablation Setup.}
We ablate two components: \textbf{(1) self-critic refinement}, \textbf{(2) reflective evidence extraction} and \textbf{(3) batching and early stopping} using the \method{} (Gemini) variant from App.~\ref{appd:settings}, unless otherwise noted.

\begin{table}[t]
\caption{
Ablation on PubMedQA with \method{} (Gemini): impact of self-critic module. \textit{Evidence-Grounded Response Rate (EGR)}: fraction of questions whose final response cites $\geq 1$ extracted findings that pass coverage and alignment checks in reflective stage.
}
\centering
\small
\setlength\tabcolsep{10pt}
\begin{tabular}{lcc}
\toprule
Method / Config. & Accuracy & EGR \\
\midrule
PubMed Reasoner    & 77.26\% & 82.64\% \\
~\textit{w/o} self-critic    & 75.60\% & 64.46\% \\
\bottomrule
\end{tabular}
\label{tab:sc_ablation}
\end{table}

\noindent\textbf{Self-Critic Refinement.} Table~\ref{tab:sc_ablation} reports accuracy and the Evidence-Grounded Response Rate (EGR)—the fraction of questions whose final response cites at least one extracted finding that passes coverage and alignment checks. Removing the self-critic markedly lowers EGR (drop of 28.20\%) and also reduces accuracy 2.2\%. This confirms that the self-critic is not only improving final correctness, but also producing robust, well-formed queries that retrieve on-point evidence.

\noindent\textbf{Reflective Evidence Extraction.}
We disable the alignment filter and instead summarize all retrieved articles before producing a final answer. As shown in Table~\ref{tab:rr_ablation}, accuracy remains unchanged, but the quality of explanations shifts: reflective integration improves reasoning soundness and clinical relevance, while introducing more diverse evidence slightly lowers grounding precision. These trends match the more fine-grained analysis in App.~\ref{app:ablation-ri}.
\noindent\textbf{Batching and Early Stopping.} The ablation study results in App.~\ref{app:ablation-es} highlight the importance of providing evidence at the appropriate granularity. High-level condensations, such as abstracts or naive summaries fail to capture essential reasoning cues, whereas our selective early-stopping framework preserves critical information and yields markedly improved performance.

\begin{table}[t]
\caption{
Ablation on PubMedQA with \method{} (Gemini): impact of reflective retrieval (RR).}
\small
\centering
\setlength\tabcolsep{10pt}
\begin{tabular}{lcc}
\toprule
Metric & \textit{w/o} RR & Ours \\
\midrule
Accuracy (\%)   & 77.24 & \textbf{77.26} \\
Reasoning Soundness     & 3.422 & \textbf{3.554}\\
Evidence Grounding     & \textbf{3.554} & 3.427 \\
Clinical Relevance     & 3.443 & \textbf{3.573} \\
Trustworthiness     & \textbf{3.432} & \textbf{3.432} \\
\bottomrule
\end{tabular}
\label{tab:rr_ablation}
\end{table}

\section{Related Work}

\noindent\textbf{LLMs \& Retrieval-Augmented Methods.} 
Despite the remarkable progress achieved by LLMs in natural language understanding, reasoning, and generation, when applied to high-stakes biomedical tasks, they often hallucinate facts or rely on outdated parametric memory, raising concerns about reliability and safety \cite{guan2023language, xu2024hallucination}. To address these issues, RAG has emerged as a promising paradigm. Existing RAG approaches either incorporate structured resources such as knowledge graphs \cite{abu2024knowledge} or augment prompts with retrieved content from domain-specific corpora \cite{arslan2024survey}. While these strategies improve factual grounding, they face persistent challenges: retrieval systems often struggle with the \textit{coverage--relevance trade-off}, returning either too little evidence or overwhelming the model with irrelevant content \cite{liu2024information}. Moreover, once an initial query is issued, most RAG pipelines lack mechanisms for iterative refinement, making them brittle in complex biomedical scenarios \cite{dai2024bias}.

\noindent\textbf{Search Query Optimization.} 
A complementary direction focuses on enhancing the query quality. 
Early work in information retrieval explored query expansion and relevance feedback \cite{sparck1974automatic, crane1951indexing}, but such methods often relied on manual heuristics \cite{arasu2001searching} and lacked semantic explanation capabilities \cite{boytsov2011indexing}. 
More recently, query optimization has also been framed as a reinforcement learning problem, where the model learns to improve retrieval performance through policy gradients or preference-based objectives such as PPO~\cite{schulman2017proximal}, DPO~\cite{rafailov2023direct}, or GRPO~\cite{shao2024deepseekmath}. 
While effective, these methods typically require a separate reward model or explicit training signals, making them computationally expensive and less flexible in specialized domains like biomedicine.
In contrast, our work introduces a training-free self-critic mechanism that performs query refinement without external supervision or gradient updates. 
Unlike prior RL-based or heuristic based method, the self-critic provides fine-grained feedback on MeSH query terms, iteratively improves retrieval quality,  mitigates error propagation and enhances factual grounding.

\noindent\textbf{Self-Reflection and Reasoning Agents.} 
Another line of research seeks to improve LLM reliability through self-reflection and agent-based reasoning. Self-reflection methods allow models to re-examine their own outputs and refine answers, while reward modeling \cite{leike2018scalable, choudhury2025process} and verbal reinforcement learning \cite{shinn2023reflexion} aim to align reasoning with human-like preferences. Self-consistency sampling further increases robustness by aggregating multiple reasoning trajectories \cite{wang2022self}. However, these methods generally intervene at the \textit{answer-generation stage}, which makes them computationally expensive and unable to prevent low-quality retrieval from propagating downstream. As a result, they provide limited control over the evidence collection process itself.

% \noindent\textbf{Query Refinement and Search Agents.} 
% A complementary direction focuses on enhancing the search process itself. Early work in information retrieval explored query expansion and relevance feedback \cite{sparck1974automatic, crane1951indexing}, but such methods often relied on manual heuristics \cite{arasu2001searching} and lacked semantic explanation capabilities \cite{boytsov2011indexing}. More recently, agent-based approaches \cite{xu2025comprehensive} have been applied to search, where an LLM iteratively issues retrieval queries and incorporates feedback from intermediate results \cite{renze2024self, xi2025survey}. While promising, most of these methods evaluate retrieval only at the document or answer level, without providing fine-grained, structured feedback on individual query terms. In the biomedical domain, where queries are often expressed in controlled vocabularies such as MeSH \cite{yu2025medreseacher}, the ability to iteratively refine queries based on coverage, alignment, and redundancy remains underexplored. Our work bridges this gap by introducing a \textbf{self-critic mechanism} that directly evaluates and improves queries during the search stage, thereby preventing error propagation and enhancing factual grounding in biomedical QA.

\section{Conclusion}
Altogether, these results demonstrate that combining structured self-critique with evidence-based integration moves biomedical QA closer to expert-level explanation, while remaining efficient and reproducible. More broadly, our findings suggest design principles for multi-stage LLM agents in high-stakes domains: shifting reflection earlier in the pipeline can prevent compounding errors; explicit grounding in external evidence improves transparency and reliability; and adaptive mechanisms such as early stopping enable practical deployment without sacrificing rigor. Notably, self-critic is especially effective in multi-step reasoning settings, where revising only problematic steps rather than regenerating the entire chain ensures both efficiency and logical consistency. These mechanisms can be generalized beyond biomedicine, providing a blueprint for building trustworthy, domain-specialized LLM systems in areas such as law, finance, and scientific discovery.
% Bibliography entries for the entire Anthology, followed by custom entries
%\bibliography{anthology,custom}
% Custom bibliography entries only
\newpage
\section{Limitations}
Although the self-critic mechanism substantially improves query quality, it remains heuristic and may inherit biases from the underlying LLM backbone, particularly in subdomains with limited training data. Additionally, the reflective retrieval stage may overlook lower-ranked yet relevant articles, potentially reducing evidence coverage in long-tail biomedical topics. Moreover, the current pipeline is largely linear at the stage-level: once reasoning progresses to later stages, earlier retrieval or query decisions cannot be revisited. This lack of backward adaptivity limits the system's ability to iteratively correct upstream errors or expand underrepresented evidence when inconsistencies arise in the final response. 
Revisiting previous stages could improve robustness and coverage but would introduce additional computational and token costs, highlighting the trade-off between accuracy and efficiency inherent to our design.

\bibliography{reference}

\appendix

% \section{Appendix}
\label{sec:appendix}
\section{Algorithms}\label{sec:algos}

Algorithms~\ref{algo:self-critic} and \ref{algo:reflective} respectively describe the steps involved in \textbf{Self-Critic Query Refinement with Query History} and \textbf{Reflective Article Retrieval with Early Stopping}.

\begin{algorithm}[tbp!]
\caption{Self-Critic Query Refinement with Query History}
\begin{algorithmic}[1]
\STATE \textbf{Input:} Initial query $q_0$ from Eq.~\ref{equ:query_gen}
\STATE \textbf{Output:} Refined query $q^*$

\STATE Initialize history $\mathcal{H} \leftarrow [q_0]$

\FOR{iteration $t = 1, \dots, T$}
    \STATE  Retrieve article metadata $\mathcal{S}_t$ (Eqs.~\ref{equ:search}--\ref{equ:metadata})
    
    \STATE Self-critic evaluation of MeSH terms:
    \STATE \quad Evaluate \textbf{coverage} (Eq.~\ref{equ:coverage}), \textbf{alignment} (Eq.~\ref{equ:alignment}), and \textbf{redundancy} (Eq.~\ref{equ:redundancy}) from $S_t$
    
    \STATE Update MeSH terms (Eq.~\ref{equ:update})
    
    \STATE Refine query $q_t$ (Eq.~\ref{equ:refine-draft})
    
    \STATE Rule-based normalization (Eq.~\ref{equ:refine-validate})
    
    \STATE Append $q_t$ to history: $\mathcal{H} \leftarrow \mathcal{H} \cup \{q_t\}$
    
    \IF{retrieval quality converges or $t = T$}
        \STATE $q^* \leftarrow q_t$; \textbf{break}
    \ENDIF
\ENDFOR

\RETURN $q^*$
\end{algorithmic}
\label{algo:self-critic}
\end{algorithm}

\begin{algorithm}[tb]
\caption{Reflective Article Retrieval with Early Stopping}
\begin{algorithmic}[1]
\STATE \textbf{Input:} Ranked result set $\mathcal{A}^\star$ and metadata $\mathcal{S}^\star$, question $Q$, maximum ranked-article budget $M_{\max}$, batch size $m$, token budget $T$
\STATE \textbf{Output:} Final evidence pool $\mathcal{E}^{(\mathrm{final})}$

\STATE Initialize filtered ranked set $\mathcal{A}^{+} \leftarrow \varnothing$
\STATE Initialize evidence pool $\mathcal{E}^{(0)} \leftarrow \varnothing$
\STATE Initialize token usage counter $b_{\text{used}} \leftarrow 0$

\STATE \textit{// Coarse filtering up to ranked-article budget}
\FOR{$i = 1$ \TO $\min\big(|\mathcal{A}^\star|,\, M_{\max}\big)$}
    \STATE Coarse filtering each article (Eq.~\ref{equ:filter})
    \IF{$v_i = \text{Yes}$}
        \STATE $\mathcal{A}^{+} \leftarrow \mathcal{A}^{+} \cup \{a_i\}$ \COMMENT{Preserves original ranking order}
    \ENDIF
\ENDFOR

\STATE Partition $\mathcal{A}^{+}$ into batches $\{\mathcal{B}_1, \dots, \mathcal{B}_K\}$ of size at most $m$

\FOR{batch index $b = 1, \dots, K$}
    \STATE \textit{// Reflective evidence extraction for $\mathcal{B}_b$}
    \STATE Extract candidate evidence (Eq.~\ref{equ:extract})
    \STATE Update evidence pool (Eq.~\ref{equ:pool})

    \STATE \textit{// Update token usage}
    \STATE $t_{\text{used}} \leftarrow t_{\text{used}} + \mathrm{TokensUsed}(\mathcal{B}_b)$
    
    \STATE Check stopping criteria (Eq.~\ref{equ:coveragecheck})
    \IF{$\text{is\_sufficient}$}
        \STATE \textbf{break} \textit{// Coverage sufficient}
    \ENDIF
    
    \IF{$t_{\text{used}} \geq T$}
        \STATE \textbf{break} \textit{// Token budget reached}
    \ENDIF
\ENDFOR

\STATE $\mathcal{E}^{(\mathrm{final})} \leftarrow \mathcal{E}^{(b)}$

\RETURN $\mathcal{E}^{(\mathrm{final})}$
\end{algorithmic}
\label{algo:reflective}
\end{algorithm}

\section{Evaluation Metrics}
\label{sec:evalmetrics}
This appendix provides detailed definitions and justifications for the evaluation metrics used in our experiments, which are designed to assess not only predictive correctness but also reasoning quality, evidence usage, and computational efficiency.

\noindent\textbf{Accuracy.}
Prediction accuracy is defined as the proportion of questions for which the model produces the correct answer label. This metric captures the model’s ability to arrive at the correct final decision and serves as a primary indicator of task performance. Accuracy is reported over the full evaluation set and is used to ensure comparability with prior biomedical question answering benchmarks.

\noindent\textbf{Explanation Quality.}
Beyond answer correctness, we evaluate the quality of model-generated explanations using an LLM-as-a-judge framework. We adopt the HealthBench evaluation protocol \cite{arora2025healthbench}, adapting the prompts to our dataset (the full prompt is provided in App.~\ref{app:llm-judge}). For each question, model outputs are assessed along four complementary axes:
\begin{itemize}[noitemsep]
    \item \textbf{Reasoning Soundness:} Whether the explanation is logically coherent, internally consistent, and free of contradictions.
    \item \textbf{Evidence Grounding:} Whether factual claims are supported by retrieved biomedical evidence, with minimal hallucination or unsupported assertions.
    \item \textbf{Clinical Relevance:} Whether the explanation directly addresses the biomedical or clinical aspects of the question in an appropriate and meaningful manner.
    \item \textbf{Trustworthiness:} Whether the response aligns with established biomedical knowledge and avoids misleading or unsafe conclusions.
\end{itemize}

Each axis is scored on a five-point Likert scale, with higher scores indicating better explanation quality. To isolate explanation quality from answer correctness, we evaluate only instances where both models predict the correct label. Pairwise comparisons are performed with randomized response order, and ties are permitted to reduce positional bias. An independent LLM (GPT) judge is used to avoid self-evaluation effects.

\noindent\textbf{Cost Analysis.}
To quantify computational efficiency, we measure four complementary cost indicators: (1) \textbf{Input token usage}, (2) \textbf{Output token usage}, (3) \textbf{Number of LLM API calls}, and (4)
\textbf{Number of PubMed search calls} issued during retrieval.
These metrics capture both the direct inference cost associated with language model usage and the operational overhead introduced by iterative retrieval and refinement. Reporting these indicators allows us to assess whether performance gains are achieved in a resource-efficient manner, which is particularly important for large-scale or deployment-oriented biomedical applications.
\noindent \textbf{Query Term Quality:}
To evaluate the effectiveness of search query formulation, we compare the MeSH terms generated by each model against the ground-truth MeSH annotations associated with each question. Precision measures the alignment of proposed terms with the question intent, while recall reflects coverage of all key biomedical concepts. This metric assesses how effectively the model identifies relevant search concepts prior to retrieval, which is crucial for downstream reasoning performance.

\noindent\textbf{Evidence Sufficiency Depth (ESD).}
We introduce \textit{Evidence Sufficiency Depth (ESD)} to characterize retrieval efficiency under early stopping. ESD is defined as the number of retrieved articles processed before the early-stopping criterion is satisfied. Since PubMed returns results in relevance-ranked order, lower ESD values indicate that sufficient supporting evidence is identified earlier in the ranked list. This metric directly reflects the model’s ability to rapidly locate adequate evidence while minimizing unnecessary paper retrieval and token consumption.

\section{\method{} Configurations}
\label{appd:settings}
\noindent \textbf{Backbone Model.} We instantiate \method{} using three model families. \method{} (GPT) employs GPT-4o with temperature $t{=}0$, which promotes faithful extraction and reduces the likelihood of introducing unsupported content that could compromise evidence fidelity. \method{} (Gemini) is a hybrid of Gemini-2.5 Pro ($t{=}0.8$) and Gemini-2.5 Flash ($t{=}0$): Pro is used primarily for the self-critic refinement and reflective extraction steps, while Flash is used for efficiency-critical calls. \method{} (Qwen) uses three Qwen variants, namely Qwen2.5-1.5B, Qwen2.5-7B and Qwen2.5-72B, all with $t{=}0$, encouraging faithful extraction and minimizing added content that could compromise evidence fidelity.

\noindent \textbf{Reflective Article Retrieval with
Early Stopping.} In the reflective stage (Sec.~\ref{sec:early-stop}), we process retrieved articles in batches of size $m{=}5$ to balance context length with reflective efficiency. To further control token cost, we impose a maximum budget of the top $M_{\max}=20$ ranked articles (i.e., at most four batches). This ensures that retrieval remains both efficient and grounded in the most relevant evidence returned by PubMed. We set the token budget 
$T$ to infinity during benchmark evaluation.

\section{Baseline Configuration}
\label{appd:baseline}
\noindent\textbf{LLM Baselines.}
We consider Gemini-2.5 Pro and GPT-4o as strong LLM baselines. In particular, we report the official leaderboard submission results as the GPT baseline on the PubMedQA benchmark.

\noindent\textbf{Human Performance.}
For PubMedQA, we report the human performance published on the official leaderboard, which serves as a reference upper bound. Human expert results are not available for MMLU-CK, and thus are not reported for that benchmark.

\noindent\textbf{RAG Method.}
Since no pre-constructed retrieval corpus or standardized RAG setup exists for the PubMedQA dataset, we implement a retrieval-augmented baseline by directly querying the PubMed search engine at inference time. Retrieved articles are incorporated as external knowledge to support response generation. To ensure a fair comparison, the RAG baseline uses the same query generation, article retrieval, summarization, and question-answering prompts as \method{}, which are provided in App.~\ref{app:prompts}.

\noindent\textbf{Self-Reflection Agent.}
The self-reflection baseline is allowed to issue PubMed search queries to retrieve supporting evidence. It generates multiple candidate responses and iteratively refines the final answer through an explicit reflection step. To ensure comparability, this baseline uses the same query generation, retrieval, summarization, and question-answering prompts as \method{}, with the addition of a dedicated prompt for the self-reflection stage.

\definecolor{appendixpurple}{RGB}{152, 78, 163}
\begin{tcolorbox}[breakable, colback=appendixpurple!10!white, colframe=appendixpurple!50!black, title=Self-reflection Prompt]

\textbf{System Prompt:}
You are a self-reflection agent for evidence-grounded biomedical question answering.

Your task is to identify conceptual gaps between the current answer and the verified sources, and to generate \textbf{one revised PubMed query} that targets missing or weakly supported concepts. When generating the revised query, you must avoid ineffective or repetitive  search patterns by consulting the search history.

A concept gap exists if one or more of the following conditions hold:
\begin{itemize}
    \item A key claim in the answer lacks direct support from the retrieved context.
    \item The context only partially addresses the question.
    \item The context is overly general and fails to capture critical biomedical specificity.
\end{itemize}

\textbf{Logical operators}:
\begin{itemize}
    \item Use \texttt{AND} to combine \emph{independent, parallel constraints}.  
          Every term connected by \texttt{AND} must be satisfied.
    \item Use \texttt{OR} only for \emph{similar or interchangeable concepts}.  
          When using \texttt{OR}, you must enclose the entire OR-group in parentheses, e.g.:  
          \texttt{(term1[mesh] OR term2[mesh])}.
\end{itemize}

\textbf{Requirements:}

\begin{enumerate}
 \item Use only the Boolean operator \texttt{AND} to connect terms.

    \item Field tag priority:
   \begin{enumerate}
     \item Place MeSH terms first and apply the \texttt{[mesh]} tag whenever possible.
     \item Place date ranges next, formatted as either \texttt{YYYY:YYYY[pdat]} or \texttt{YYYY/MM/DD:YYYY/MM/DD[pdat]}.
     \item Place all remaining terms last. Do not apply special field tags unless explicitly specified.
   \end{enumerate}

    \item Spacing: Separate each term and each \texttt{AND} with exactly one space.

    \item Date range format:  
   \texttt{YYYY:YYYY[pdat]} or \texttt{YYYY/MM/DD:YYYY/MM/DD[pdat]}
\end{enumerate}

\textbf{Natural Language Question:}

\{natural\_language\_question\}

\textbf{Verified Sources:}

\{verified\_sources\}

\textbf{Answer:}

\{rationale\_answer\}

\textbf{Search History (if any):}

\{search\_history\}

\textbf{Output:}

Return the answer strictly as JSON following this schema:

\begin{verbatim}
{
  "query": 
    "The final PubMed query string as 
    a single string.",
  "rationale": 
    "A brief explanation describing
    the identified concept gaps and
    how the revised query addresses
    these gaps while following the
    query construction rules."
}
\end{verbatim}

Do not add any explanation or additional text outside the JSON.

\end{tcolorbox}

\noindent
We instantiate both the RAG method and the self-reflection agent using two backbone models, GPT-4o and Gemini-2.5 Pro/Flash, following the same parameter settings as described in App.~\ref{appd:settings}. To encourage diverse reasoning trajectories, we employ higher decoding temperatures during the reflection stage (GPT at \(t = 1.0\) and Gemini at \(t = 0.8\)).

\section{Prompt Design}
\label{app:prompts}

\definecolor{appendixpurple}{RGB}{152, 78, 163}
\begin{tcolorbox}[breakable, colback=appendixpurple!10!white, colframe=appendixpurple!50!black, title=Query Generation Prompt]

\textbf{System Prompt:}

You are an expert in PubMed search syntax.

Your task is to convert the provided natural language description of the desired literature, together with any contextual information, into a single valid PubMed query string.

\textbf{Requirements:}

\begin{enumerate}
 \item Use only the Boolean operator \texttt{AND} to connect terms.

    \item Field tag priority:
   \begin{enumerate}
     \item Place MeSH terms first and apply the \texttt{[mesh]} tag whenever possible.
     \item Place date ranges next, formatted as either \texttt{YYYY:YYYY[pdat]} or \texttt{YYYY/MM/DD:YYYY/MM/DD[pdat]}.
     \item Place all remaining terms last. Do not apply special field tags unless explicitly specified.
   \end{enumerate}

    \item Spacing: Separate each term and each \texttt{AND} with exactly one space.

    \item Date range format:  
   \texttt{YYYY:YYYY[pdat]} or \texttt{YYYY/MM/DD:YYYY/MM/DD[pdat]}
\end{enumerate}

\textbf{Natural Language Question:}

\{natural\_language\_question\}

\textbf{Additional Context (if any):}

\{context\}

\textbf{Output:}

Return the answer strictly as JSON following this schema:

\begin{verbatim}
{
  "query": 
    "The final PubMed query string as 
    a single string.",
  "rationale": 
    "A brief explanation of term 
    selection, field tags, ordering, 
    use of AND, and how the context 
    was incorporated."
}
\end{verbatim}

Do not add any explanation or additional text outside the JSON.

\end{tcolorbox}

\definecolor{appendixpurple}{RGB}{152, 78, 163}
\begin{tcolorbox}[breakable, colback=appendixpurple!10!white, colframe=appendixpurple!50!black, title=Self-critic Prompt]

\textbf{System Prompt:}
You are a PubMed search planning assistant.

Your task is to produce \textbf{one improved PubMed query} for the next search step by:
\begin{itemize}
    \item Interpreting the natural-language question and any additional context.
    \item Using search history to avoid ineffective or repetitive patterns.
    \item Evolving the candidate term set using coverage, alignment, and redundancy feedback.
    \item Ensuring the final query strictly follows the requirements.
\end{itemize}

\textbf{Feedback Signals:}

If no context is provided, return -1 for the corresponding signals. In producing the improved query, you must incorporate evolving feedback from:
\begin{enumerate}
    \item \textbf{Coverage} — 1 if the provided context sufficiently represents the concepts relevant to the question; 0 otherwise.
    \item \textbf{Alignment} — 1 if the provided context is relevant and appropriately focused on the question; 0 otherwise.
    \item \textbf{Redundancy} — 1 if there are \textbf{no} overlapping, unnecessary, or logically unintended terms; 0 otherwise.
\end{enumerate}

\textbf{Logical operators}:
\begin{itemize}
    \item Use \texttt{AND} to combine \emph{independent, parallel constraints}.  
          Every term connected by \texttt{AND} must be satisfied.
    \item Use \texttt{OR} only for \emph{similar or interchangeable concepts}.  
          When using \texttt{OR}, you must enclose the entire OR-group in parentheses, e.g.:  
          \texttt{(term1[mesh] OR term2[mesh])}.
\end{itemize}

\textbf{Requirements:}

\begin{enumerate}
 \item Use only the Boolean operator \texttt{AND} to connect terms.

    \item Field tag priority:
   \begin{enumerate}
     \item Place MeSH terms first and apply the \texttt{[mesh]} tag whenever possible.
     \item Place date ranges next, formatted as either \texttt{YYYY:YYYY[pdat]} or \texttt{YYYY/MM/DD:YYYY/MM/DD[pdat]}.
     \item Place all remaining terms last. Do not apply special field tags unless explicitly specified.
   \end{enumerate}

    \item Spacing: Separate each term and each \texttt{AND} with exactly one space.

    \item Date range format:  
   \texttt{YYYY:YYYY[pdat]} or \texttt{YYYY/MM/DD:YYYY/MM/DD[pdat]}
\end{enumerate}

\textbf{Natural Language Question:}

\{natural\_language\_question\}

\textbf{Additional Context (if any):}

\{search\_meta\}

\textbf{Search History (if any):}

\{search\_history\}

\textbf{Output:}

Return the answer strictly as JSON following this schema:

\begin{verbatim}
{
  "query": 
    "The final PubMed query string as 
    a single string.",
  "rationale": 
    "A brief explanation of term 
    selection, field tags, ordering, 
    use of logical operators, 
    and how the context and feedback 
    were incorporated."
    "feedback": {
        "coverage": int,
         "coverage_suggestion": 
         "Suggested improvements",
        "alignment": int,
         "alignment_suggestion": 
         "Suggested improvements",
        "redundancy": int,
         "redundancy_suggestion": 
         "Suggested improvements"
    }
}
\end{verbatim}

Do not add any explanation or additional text outside the JSON.

\end{tcolorbox}

\definecolor{appendixpurple}{RGB}{152, 78, 163}
\begin{tcolorbox}[breakable,
                  colback=appendixpurple!10!white,
                  colframe=appendixpurple!50!black,
                  title=Reflective Article Retrieval Prompt]

\textbf{System Prompt:}

You are a reflection assistant.

Your task is to determine whether the provided search results with current context (if provided) contain enough relevant and specific information to answer the question.

\textbf{Natural Language Question:} 

\{natural\_language\_question\}

\textbf{Search Results:} 

\{search\_results\_str\}

\textbf{Additional Context (if any):}

\{context\}

\textbf{Output:}

Return your answer strictly as JSON following this schema:

\begin{verbatim}
{
  "is_sufficient": true | false,
  "rationale": "Concise explanation
  of why the information is
  sufficient or insufficient.",
  "needed_pmids": 
    ["PMID1", "PMID2", ...]  
    # PMIDs of additional relevant 
    articles, if any
}
\end{verbatim}

Do not include any explanation or text outside the JSON.

\end{tcolorbox}

\definecolor{appendixpurple}{RGB}{152, 78, 163}
\begin{tcolorbox}[breakable, colback=appendixpurple!10!white, colframe=appendixpurple!50!black, title=Summary Prompt]

\textbf{System Prompt:}

You are a professional academic rewriting assistant.  

Your task is to transform the provided raw sources into a single, semantically coherent, and well-structured paragraph.  

\textbf{Requirements:}
\begin{enumerate}
    \item Use only the information from the provided raw sources, without adding external content.
    \item Preserve all original in-text citations exactly as they appear (e.g., [PMID: xxxx]).
    \item Ensure the paragraph is logically connected, concise, and scientifically rigorous.
\end{enumerate}

\textbf{Raw Sources:}

\{raw\_sources\}

\textbf{Output:}

Return the answer strictly as JSON following this schema:

\begin{verbatim}
{
  "verified_sources": 
   "The final rewritten paragraph 
   as a single string.",
}
\end{verbatim}

Do not add any explanation or additional text outside the JSON.

\end{tcolorbox}

\definecolor{appendixpurple}{RGB}{152, 78, 163}
\begin{tcolorbox}[breakable, colback=appendixpurple!10!white, colframe=appendixpurple!50!black, title=Question Answering Prompt]

\textbf{System Prompt:}

You are an expert assistant.

When sources are provided, you should primarily base your answer on the information in the sources. 
If the sources do not contain enough information to fully answer the question, you may supplement your answer using your own knowledge.

Provide your answer and a clear rationale explaining how you arrived at it.

\textbf{Natural language question:}

\{natural\_language\_question\}

\textbf{Task instruction:} 

\{task\_instruction\}

\textbf{Additional context (if any):} 
{context}

\textbf{Sources:}
{sources}

\textbf{Output:}

Return the answer strictly as JSON following this schema:

\begin{verbatim}
{
  "answer": "Your answer according to 
  the task instruction.",
  "rationale": "A clear explanation of 
  how you arrived at the answer.
}
\end{verbatim}

Do not add any explanation or additional text outside the JSON.

\end{tcolorbox}

\section{Qwen Results}
\label{app:qwen-results}
Table~\ref{tab:qwen_results} summarizes the performance of all evaluated methods on PubMedQA and MMLU-CK. We report accuracy (\%) across three Qwen2.5 model scales (1.5B, 7B, and 72B). Results are organized by (1) base model performance without retrieval, (2) self-reflection agent, and (3) our \method{} framework.
\begin{table}[H]
\small
\centering
\caption{
Accuracy on PubMedQA and MMLU-CK test sets for \method{} Qwen2.5 variants, strong LLMs, and self-reflection agents. 
Best result per column is \textbf{bold-faced}.
}
\label{tab:qwen_results}
\setlength\tabcolsep{4pt}
\begin{tabular}{lcc}
\toprule
Model/Method & PubMedQA & MMLU-CK \\
\midrule
\multicolumn{1}{l}{\textit{Qwen2.5-1.5B}} & & \\
% \midrule
~~~Without retrieval & \textbf{68.08\%} & \textbf{55.67\%} \\
~~~Self-reflection agent & 68.06\% & 51.44\% \\
~~~\textit{\method{}} & 67.24\% & 51.42\% \\
\midrule
\multicolumn{1}{l}{\textit{Qwen2.5-7B}} & & \\
% \midrule
~~~Without retrieval & 70.20\% & \textbf{56.64\%} \\
~~~Self-reflection agent & \textbf{71.12\%} & 56.11\% \\
~~~\textit{\method{}} & 71.08\% & 56.09\% \\
\midrule
\multicolumn{1}{l}{\textit{Qwen2.5-72B}} & & \\
% \midrule
~~~Without retrieval & 74.89\% & 58.36\% \\
~~~Self-reflection agent & 76.22\% & 59.67\% \\
~~~\textit{\method{}} & \textbf{76.42\%} & \textbf{60.95\%} \\
% \multicolumn{1}{l}{\textit{Without retrieval}} & & \\
% % \midrule
% Qwen2.5-1.5B & \textbf{68.08\%} & \textbf{55.67\%} \\
% Qwen2.5-7B & 70.20\% & \textbf{56.64\%} \\
% Qwen2.5-72B & 74.89\% & 58.36\% \\
% \midrule
% \multicolumn{1}{l}{\textit{Self-reflection agent}} & & \\
% % \midrule
% ~with Qwen2.5-1.5B & 68.06\% & 51.44\% \\
% ~with Qwen2.5-7B & \textbf{71.12\%} & 56.11\% \\
% ~with Qwen2.5-72B & 76.22\% & 59.67\% \\
% \midrule
% \multicolumn{1}{l}{\textit{\method{}}} & & \\
% % \midrule
% ~with Qwen2.5-1.5B & 67.24\% & 51.42\% \\
% ~with Qwen2.5-7B & 71.08\% & 56.09\% \\
% ~with Qwen2.5-72B & \textbf{76.42\%} & \textbf{60.95\%} \\
\bottomrule
\end{tabular}
\end{table}

The effectiveness of \method{} grows with model scale: while retrieval provides limited benefit for smaller Qwen variants (1.5B and 7B), the 72B model achieves the strongest gains, outperforming both its no-retrieval and self-reflection baselines on PubMedQA and MMLU-CK.

\section{Case study}
\label{app:case}

We present several case studies comparing \method{} with GPT-4o on PubMedQA-style questions. Each case includes the user question, the model’s response, and the rationale provided. For \method{}, the rationale is grounded in biomedical literature retrieved from PubMed, with inline citations (PMID identifiers) referencing specific studies.  

These examples illustrate how \method{} improves over direct LLM reasoning in several ways: (i) grounding answers in authoritative biomedical sources, thereby reducing hallucinations; (ii) providing more detailed rationales that capture not only correlations but also evidence from longitudinal studies, randomized controlled trials, or meta-analyses; and (iii) maintaining interpretability through explicit evidence chains.

\definecolor{caseblue}{RGB}{33, 113, 181}

\newtcolorbox{casebox}[2][]{
  breakable,
  colback=caseblue!5!white,
  colframe=caseblue!75!black,
  title={\textbf{Question:} #2},
  #1
}

% Example Case
\begin{casebox}{Does early introduction of sugar-sweetened beverages increase the risk of obesity in children?}

\textbf{GPT-4o:} Yes \\
\textbf{Rationale:} Evidence consistently shows that early introduction of sugar-sweetened beverages is linked with increased caloric intake, poor dietary habits, and higher risk of obesity in children.

\vspace{6pt}

\textbf{\method{} (GPT-4o):} Yes \\
\textbf{Rationale:} Evidence from longitudinal and randomized controlled trials indicates that early introduction of juice and sugar-sweetened beverages is associated with higher intake later in childhood, a well-established obesity risk factor. 

A cohort study [PMID:37150479] found that introducing juice before 6 months led to significantly higher juice consumption at 24 months, while an RCT [PMID:36146909] showed that parental education reduced early SSB introduction and may lower obesity risk.
\end{casebox}

\begin{casebox}{Is breastfeeding associated with a lower risk of childhood obesity?}

\textbf{GPT-4o:} Yes \\
\textbf{Rationale:} Multiple studies have demonstrated that breastfeeding, especially exclusive breastfeeding, is associated with a modest but significant reduction in childhood obesity risk compared to formula feeding.

\vspace{6pt}

\textbf{\method{} (GPT-4o):} Yes \\
\textbf{Rationale:} The evidence is mixed but overall supports a protective effect. One study found no association between feeding in the first 4 months and obesity risk after adjusting for confounders [PMID:40550602]. However, metabolomics analyses link formula feeding with obesity-related metabolic profiles [PMID:40452153]. 

A large cohort study found longer breastfeeding duration reduced risk, with effects differing by sex ($\geq8$ months protective in girls, 3–7 months protective in boys) [PMID:40440349].
\end{casebox}

\begin{casebox}{Can family-based lifestyle interventions reduce obesity prevalence in children?}

\textbf{GPT-4o:} Yes \\
\textbf{Rationale:} Family-based lifestyle interventions targeting diet, physical activity, and behavior change are effective in reducing BMI and obesity prevalence in children.

\vspace{6pt}

\textbf{\method{} (GPT-4o):} Yes \\
\textbf{Rationale:} Studies show that parental obesity, education, and health behaviors strongly influence child obesity risk. Family lifestyle patterns persist over time [PMID:29807535]. Interventions targeting the whole family, including parental health behaviors and socioeconomic supports, are associated with healthier child weight outcomes [PMID:40312675, PMID:36904286].
\end{casebox}

\begin{casebox}{Is there a link between fast food consumption and childhood obesity?}

\textbf{GPT-4o:} Yes \\
\textbf{Rationale:} High frequency of fast food consumption is associated with higher caloric intake, increased fat/sugar consumption, and greater obesity risk.

\vspace{6pt}

\textbf{\method{} (GPT-4o):} Yes \\
\textbf{Rationale:} Evidence shows fast food and ultra-processed food (UPF) consumption contributes to childhood obesity. UPFs accounted for 40\% of caloric intake in one study, with fast foods alone contributing 7.8\% and being correlated with waist circumference and overweight/obesity prevalence [PMID:40102562]. 

Another large study found maternal junk food consumption increased children’s obesity odds by 39\% [PMID:40481596].
\end{casebox}

\begin{casebox}{Do school-based nutrition programs improve weight outcomes in children?}

\textbf{GPT-4o:} Yes \\
\textbf{Rationale:} School-based nutrition programs promote healthier eating and physical activity, leading to modest improvements in BMI and obesity outcomes.

\vspace{6pt}

\textbf{\method{} (GPT-4o):} Yes \\
\textbf{Rationale:} A 3-year prospective cohort study in southern Taiwan found that lunchtime education and class-based nutrition programs increased normal-weight prevalence and reduced overweight/obesity. The BMI trajectory flattened compared to regional trends, approaching national standards [PMID:32381350].
\end{casebox}

\begin{casebox}{Are children with sleep deprivation at higher risk for obesity?}

\textbf{GPT-4o:} Yes \\
\textbf{Rationale:} Sleep deprivation alters appetite-regulating hormones, increases snacking, and is consistently linked with higher obesity risk in children.

\vspace{6pt}

\textbf{\method{} (GPT-4o):} Yes \\
\textbf{Rationale:} Epidemiological studies link insufficient sleep to adverse metabolic, endocrine, and energy balance changes leading to obesity [PMID:38012858, PMID:38004130]. 

A large cohort found insufficient sleep increased early maturation risk in girls, with obesity as a mediator [PMID:38065690].
\end{casebox}

\begin{casebox}{Is there an association between sedentary behavior and pediatric obesity?}

\textbf{GPT-4o:} Yes \\
\textbf{Rationale:} Prolonged sitting and excessive screen time are strongly associated with higher adiposity and obesity in children.

\vspace{6pt}

\textbf{\method{} (GPT-4o):} Yes \\
\textbf{Rationale:} Studies show sedentary behavior is linked with pediatric obesity. Preschool children with more sedentary time and screen use had greater odds of overweight/obesity [PMID:40751144]. 

Among Chinese adolescents, screen-based sedentary behavior correlated with higher BMI, mediated by weight concerns [PMID:40611060]. 

Reviews confirm reducing sedentary behavior improves weight outcomes [PMID:40494564].
\end{casebox}

\begin{casebox}{Does screen time reduction improve BMI in obese children?}

\textbf{GPT-4o:} Yes \\
\textbf{Rationale:} Reducing screen time decreases sedentary behavior, encourages activity, and can improve BMI outcomes.

\vspace{6pt}

\textbf{\method{} (GPT-4o):} No \\
\textbf{Rationale:} A meta-analysis of 14 RCTs (n=1,894) found screen time interventions reduced weekly screen use but did not significantly improve BMI or BMI z-scores. While effective for reducing sedentary behavior, they are insufficient alone for BMI improvement [PMID:35248682].
\end{casebox}

\begin{casebox}{Does physical activity combined with nutrition counseling improve BMI in obese children?}

\textbf{GPT-4o:} Yes \\
\textbf{Rationale:} Combining physical activity with nutrition counseling effectively reduces BMI in obese children.

\vspace{6pt}

\textbf{\method{} (GPT-4o):} Yes \\
\textbf{Rationale:} A Cochrane meta-analysis [PMID:28639319] found multi-component interventions (diet, activity, behavior change) significantly reduced BMI ($\approx$ -0.53 kg/m²). 

Another meta-analysis [PMID:32408483] reported that only combined interventions produced significant reductions ($\approx$ -0.30, p<0.001), whereas exercise alone did not. Individual RCTs confirm these effects.
\end{casebox}

\section{Prompt: LLM-as-Judge}
\label{app:llm-judge}
The four dimensions capture complementary aspects of reasoning quality in biomedical QA:

\begin{itemize}
    \item \textbf{Reasoning Soundness} focuses on the logical structure of the answer. A response should be coherent, free of contradictions, and follow a clear reasoning chain. Even if evidence is present, poor logical flow reduces interpretability.
    \item \textbf{Evidence Grounding} emphasizes factual support. Claims must be explicitly backed by biomedical literature, reducing hallucinations. An answer with strong logical form but unsupported claims will score lower here.
    \item \textbf{Clinical Relevance} measures whether the reasoning directly addresses the biomedical question in an evidence-based manner. Answers that drift into tangential findings or generic biomedical facts are penalized, even if factually correct.
    \item \textbf{Trustworthiness} assesses safety and adherence to established biomedical knowledge. This ensures that the reasoning avoids misleading or potentially harmful statements, which is critical in high-stakes clinical contexts.
\end{itemize}

Together, these criteria provide a multidimensional evaluation: \textbf{soundness} ensures logical clarity, \textbf{grounding} ensures factual reliability, \textbf{relevance} ensures task alignment, and \textbf{trustworthiness} ensures clinical safety. By jointly considering all four, we obtain a more faithful measure of reasoning quality than predictive accuracy alone.

\definecolor{appendixpurple}{RGB}{152, 78, 163}
\begin{tcolorbox}[breakable,colback=appendixpurple!10!white, colframe=appendixpurple!50!black, title=LLM-as-Judge Prompt]
\textbf{System Prompt:} You are a neutral medical evaluator. Compare two answers from medical language models
for a PubMedQA-style question. Judge *reasoning quality only* (not model identity).

Question:
natural\_language\_question

Answer A:
answer\_a

Answer B:
answer\_b

Evaluate each answer independently on four dimensions (1-5):

1) Reasoning Soundness - logical, coherent, internally consistent.

2) Evidence Grounding - claims supported by biomedical evidence; no hallucinations.

3) Clinical Relevance - directly addresses the question in an evidence-based manner.

4) Trustworthiness - safe, conforms to biomedical knowledge; not misleading.

Instructions:

- Assign a numeric score (1-5) for each dimension to both A and B.

- Give a brief justification (less than 2 sentences) for each score.

- Provide an overall verdict based on reasoning quality: "A", "B", or "tie".

- Do not mention model names or speculate on sources.

- Output strictly valid JSON matching this schema (and nothing else):

\begin{verbatim}
{
  "Answer A": {
    "Reasoning Soundness": {
      "score": <int>,
      "justification": "<string>"
    },
    "Evidence Grounding": {
      "score": <int>,
      "justification": "<string>"
    },
    "Clinical Relevance": {
      "score": <int>,
      "justification": "<string>"
    },
    "Trustworthiness": {
      "score": <int>,
      "justification": "<string>"
    }
  },
  "Answer B": {
    "Reasoning Soundness": {
      "score": <int>,
      "justification": "<string>"
    },
    "Evidence Grounding": {
      "score": <int>,
      "justification": "<string>"
    },
    "Clinical Relevance": {
      "score": <int>,
      "justification": "<string>"
    },
    "Trustworthiness": {
      "score": <int>,
      "justification": "<string>"
    }
  },
}
\end{verbatim}

\end{tcolorbox}

\section{Evidence Sufficiency Depth}\label{app:esd}
\begin{figure}[t]
\centering 
\includegraphics[width=\linewidth]{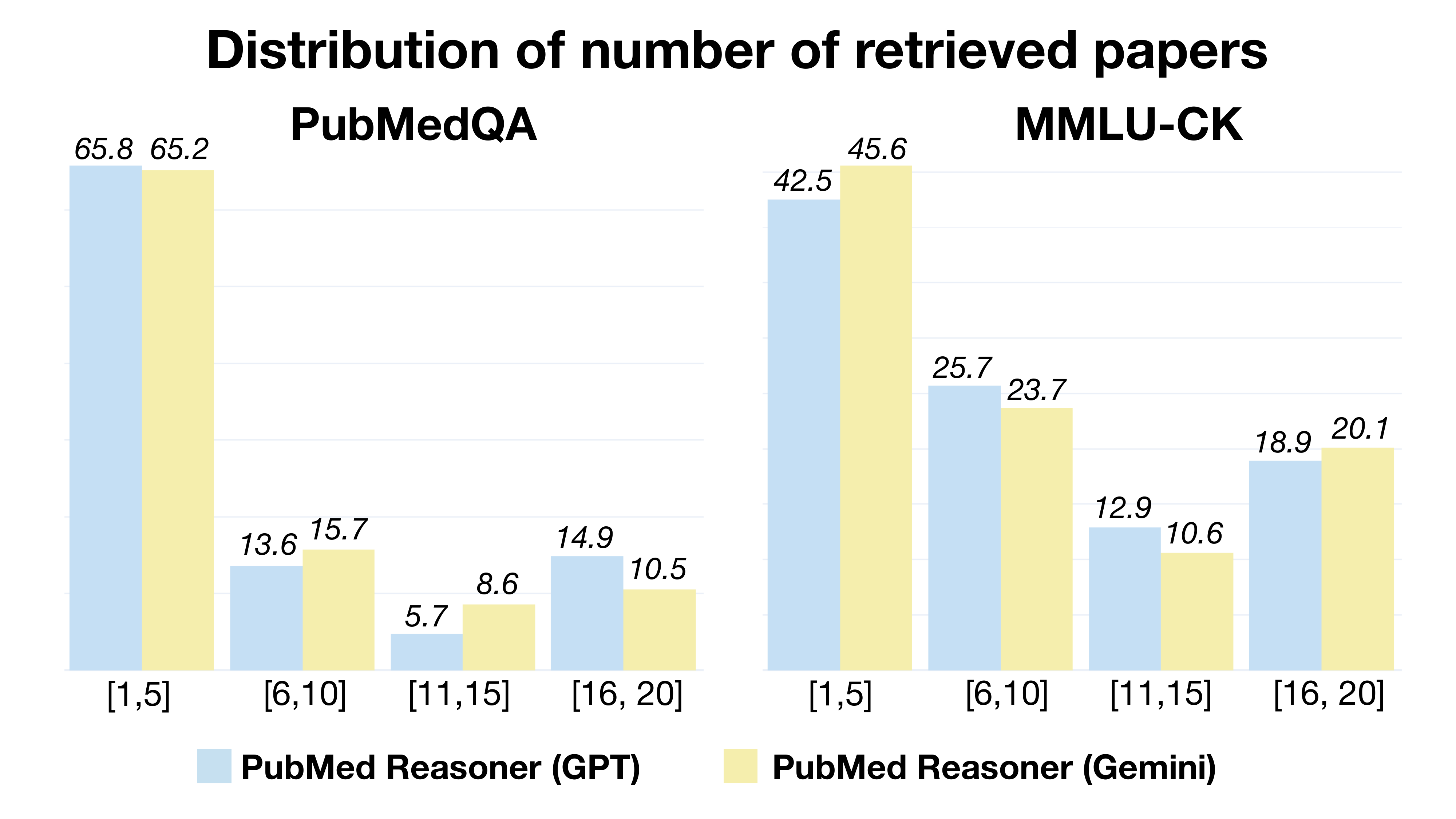}
\caption{
Effectiveness of the reflective integration stage. Distribution of retrieval depth before early stopping.
}
\label{fig:precent}
\end{figure}

Fig.~\ref{fig:precent} illustrates the distribution of the number of retrieved papers required before reaching evidence sufficiency across PubMedQA and MMLU-CK datasets.

\noindent \textbf{PubMedQA.}  Most questions achieve evidence sufficiency within the first five retrieved articles (65.8\% for GPT and 65.2\% for Gemini). Over 80\% require no more than ten articles, demonstrating that \method{} efficiently capitalizes on PubMed’s ranking to identify relevant studies early.
 
\noindent \textbf{MMLU-CK.}
Although questions are more diverse and less structured, more than 70\% of cases still reach sufficiency within ten articles. This indicates that the reflective retriever generalizes well beyond domain-specific datasets.
Fig.~\ref{fig:precent} illustrates the distribution of the number of retrieved papers required before reaching evidence sufficiency across PubMedQA and MMLU-CK datasets. 
For both backbones (GPT and Gemini), the majority of questions obtain sufficient supporting evidence within the first five retrieved articles—65.8\% for PubMed Reasoner (GPT) and 65.2\% for PubMed Reasoner (Gemini). 

Across both datasets, ESD highlights \method{}'s ability to rapidly and adaptively locate adequate evidence, minimizing unnecessary retrieval while preserving grounding quality.

\section{Qualitative Error Analysis and Challenging Cases}
\label{app:error-analysis}
To provide deeper insight into the limitations of \method{}, we include here a qualitative error taxonomy alongside representative failure cases. Our analysis is based on mispredicted examples from the PubMedQA dataset using the \method{} (GPT) variant.

\noindent\textbf{Error Taxonomy and Distribution.}
We categorize observed failures into three common patterns. The distribution of error types is as follows:
\begin{itemize}
    \item \textbf{Case 1: No articles retrieved (54.1\%)} \\
    Often caused by overly specific MeSH terms added during refinement, leading the query to become too restrictive.
    \item \textbf{Case 2: Incomplete MeSH coverage (29.3\%)} \\
    The refined query omits critical concepts necessary for retrieving the correct evidence.
    \item \textbf{Case 3: Other errors (16.6\%)} \\
    Includes request timeouts, API rate limits, and unexpected server or connection errors.
\end{itemize}

\vspace{0.5em}
\noindent\textbf{Case 1 Example: No Articles Retrieved.}
\begin{itemize}
    \item \textbf{Question:} Do molecular signatures of mood stabilisers highlight the role of the transcription factor REST/NRSF?
    \item \textbf{Ground-truth MeSH terms:} \\
    Antimanic Agents AND Cell Line AND Humans AND Lithium Compounds AND Repressor Proteins AND Transcriptome AND Valproic Acid
    \item \textbf{Refined query MeSH terms:} \\
    Neuroblastoma AND Cells, Cultured AND Cell Line AND Lithium AND \textbf{RE1-Silencing Transcription Factor} AND Valproic Acid AND \textbf{GAD1} AND Humans AND Time Factors
\end{itemize}
\noindent\textit{Overly specific terms (bold) caused the search to become too narrow, retrieving zero results.}

\vspace{0.75em}
\noindent\textbf{Case 2 Example: Incomplete MeSH Coverage.}
\begin{itemize}
    \item \textbf{Question:} Does blood viscosity but not shear stress associate with delayed flow-mediated dilation?
    \item \textbf{Ground-truth MeSH terms:} \\
    Age Factors AND Aged AND Blood Viscosity AND Brachial Artery AND Female AND Humans AND Male AND Middle Aged AND Stress, Mechanical AND Vasodilation
    \item \textbf{Refined query MeSH terms:} \\
    Middle Aged AND Aged AND Vasodilation AND Female AND Humans AND Male AND Brachial Artery AND Blood Flow Velocity AND Regression Analysis AND Cross-Sectional Studies
\end{itemize}
\noindent\textit{Critical concepts such as \textbf{Blood Viscosity} and \textbf{Stress, Mechanical} were omitted, resulting in incomplete retrieval.}

\vspace{0.75em}
\noindent\textbf{Case 3 Example: System or API Errors.}
\noindent Representative failures include rate limits, connection issues, and unexpected HTTP errors.

Together, these examples highlight both conceptual and infrastructural failure modes, illustrating where \method{} can be further improved in robustness, coverage, and query generalization.

\section{Ablation Study for Reflective Evidence Extraction}
\label{app:ablation-ri}
Table~\ref{abl:add} presents the ablation results for the reflective integration stage on the PubMedQA test set. 
The comparison between \method{} and its variant without reflective retrieval reveals that this stage substantially improves overall reasoning quality. 
In terms of win/tie/loss rates judged by GPT-4.1, \method{} achieves clear advantages in \textit{Reasoning Soundness} (76.4\% wins) and \textit{Clinical Relevance} (58.6\% wins), indicating that the reflective process helps the model produce more coherent and question-aligned explanations. 
The improvement is also reflected in the corresponding average Likert scores, which increase from 3.422 to 3.554 for soundness and from 3.443 to 3.573 for relevance. Interestingly, while the score for \textit{Evidence Grounding} slightly declines (from 3.554 to 3.427), this pattern aligns with our earlier observation that the reflective stage introduces a broader range of evidence, sometimes increasing conceptual diversity at the cost of strict grounding precision. 
Nevertheless, the consistent or improved performance across other metrics demonstrates that reflective integration enhances interpretability and contextual reasoning, enabling \method{} to deliver explanations that are both clinically meaningful and logically sound.

\begin{table}[t!]
\caption{
Explanation quality evaluation on PubMedQA test sets. 
Left: win/tie/loss rates from pairwise comparisons judged by GPT-4.1. 
Right: average Likert scores (1–5). 
We compare \method{} with and without the reflective retrieval stage.
}
\centering
\small
\setlength{\tabcolsep}{6pt}
\begin{tabular}{lccccc}
\toprule
\multirow{2}{*}{Metric}
 & \multicolumn{3}{c}{Loss / Tie / Win (\%)} 
 & \multicolumn{2}{c}{Avg. Likert (1–5)} \\
\cmidrule(lr){2-4} \cmidrule(lr){5-6}
 & w/o & Tie & Ours & w/o & Ours \\
\midrule
\begin{tabular}[c]{@{}l@{}}Reasoning \\ Soundness\end{tabular}
  & 19.6 & 4.0 & \textbf{76.4} & 3.422 & \textbf{3.554} \\
\begin{tabular}[c]{@{}l@{}}Evidence \\ Grounding\end{tabular}
  & \textbf{26.2} & 54.3 & 19.4 & \textbf{3.554} & 3.427 \\
\begin{tabular}[c]{@{}l@{}}Clinical \\ Relevance\end{tabular}
  & 17.6 & 23.8 & \textbf{58.6} & 3.443 & \textbf{3.573} \\
\begin{tabular}[c]{@{}l@{}}Trust- \\ worthiness\end{tabular}
  & 10.0 & 75.8 & \textbf{14.2} & \textbf{3.432} & \textbf{3.432} \\
\bottomrule
\end{tabular}
\label{abl:add}
\end{table}

\section{Ablation Study on Early-Stopping Mechanism}
\label{app:ablation-es}

To further investigate the effectiveness of the early-stopping mechanism, we conducted an ablation study on the PubMedQA dataset Gemini variant. In our setup, the model processes up to 20 retrieved papers, reading them in batches of five. The early-stopping mechanism terminates the reading process once the model determines that the available context is sufficient to answer the question.

To assess the performance implications of omitting this mechanism, we evaluated alternative settings in which the top 20 retrieved papers were supplied in condensed form. Owing to context-length constraints, we could not provide the full text of all 20 papers simultaneously. Instead, we considered two reduced-context variants:  
\textbf{(1) providing only the abstract (metadata) of each paper}, and  
\textbf{(2) generating a per-paper summary and concatenating these summaries} before the final question-answering step.

Table~\ref{tab:es-ablation} reports the performance of each configuration. Both reduced-context approaches resulted in substantial degradation, demonstrating that the absence of key extracted observations from each paper significantly impairs reasoning quality.

\begin{table}[H]
\centering
\begin{tabular}{lc}
\toprule
\textbf{Method} & \textbf{Accuracy (\%)} \\
\midrule
Abstract Only & 67.20 \\
Paper Summary & 64.12 \\
\method{} & \textbf{77.26} \\
\bottomrule
\end{tabular}
\caption{Ablation study comparing reduced-context variants to our selective early-stopping framework on PubMedQA.}
\label{tab:es-ablation}
\end{table}

These results highlight the importance of providing evidence at the appropriate granularity. High-level condensations, such as abstracts or naive summaries fail to capture essential reasoning cues, whereas our selective early-stopping framework preserves critical information and yields markedly improved performance.

\end{document}